# Teleo-Reactive Programs for Agent Control

**Nils J. Nilsson** NILSSON@CS.STANFORD.EDU
*Robotics Laboratory, Department of Computer Science*
*Stanford University, Stanford, CA 94305 USA*

## Abstract

A formalism is presented for computing and organizing actions for autonomous agents in dynamic environments. We introduce the notion of *teleo-reactive (T-R) programs* whose execution entails the construction of circuitry for the continuous computation of the parameters and conditions on which agent action is based. In addition to continuous feedback, T-R programs support parameter binding and recursion. A primary difference between T-R programs and many other circuit-based systems is that the circuitry of T-R programs is more compact; it is constructed at run time and thus does not have to anticipate *all* the contingencies that might arise over all possible runs. In addition, T-R programs are intuitive and easy to write and are written in a form that is compatible with automatic planning and learning methods. We briefly describe some experimental applications of T-R programs in the control of simulated and actual mobile robots.

## 1. Introduction

Autonomous agents, such as mobile robots, typically operate in dynamic and uncertain environments. Such environments can be sensed only imperfectly, effects on them are not always completely predictable, and they may be subject to changes not under the agent's control. Designing agents to operate in these environments has presented challenges to the standard methods of artificial intelligence, which are based on explicit declarative representations and reasoning processes. Prominent among the alternative approaches are the so-called *behavior-based*, *situated*, and *animat* methods (Brooks, 1986; Maes, 1989; Kaelbling & Rosenschein, 1990; Wilson, 1991), which convert sensory inputs into actions in a much more direct fashion than do AI systems based on representation and reasoning. Many of these alternative approaches share with control theory the central notion that continuous feedback from the environment is a necessary component of effective action.

Perhaps it is relatively easier for control theorists than it is for computer scientists to deal with continuous feedback because control theorists are accustomed to thinking of their controlling mechanisms as composed of analog electrical circuits or other physical systems rather than as automata with discrete read-compute-write cycles. The notions of goal-seeking servo-mechanisms, homeostasis, feedback, filtering, and stability—so essential to control in dynamic environments—were all developed with analog circuitry in mind. Circuits, by their nature, are continously responsive to their inputs.

In contrast, some of the central ideas of computer science, namely sequences, events, discrete actions, and subroutines, seem at odds with the notion of continuous feedback. For example, in conventional programming when one program calls another, the calling program is suspended until the called program returns control. This feature is awkward in applications in which the called program might encounter unexpected environmental circumstances with which it was not designed to cope. In such cases, the calling program can regain control only through interrupts explicitly provided by the programmer.

To be sure, there have been attempts to blend control theory and computer science. For example, the work of Ramadge and Wonham (Ramadge & Wonham, 1989) on *discrete-event systems* has used the computer science notions of events, grammars, and discrete states to study the control of processes for which those ideas are appropriate. A book by Dean and Wellman (Dean & Wellman, 1991) focusses on the overlap between control theory and artificial intelligence. But there has been little effort to import fundamental control-theory ideas into computer science. That is precisely what I set out to do in this paper.

I propose a computational system that works differently than do conventional ones. The formalism has what I call *circuit semantics* (Nilsson, 1992); program execution produces (at least conceptually) electrical circuits, and it is these circuits that are used for control. While importing the control-theory concept of continuous feedback, I nevertheless want to retain useful ideas of computer science. My control programs will have parameters that can be bound at run time and passed to subordinate routines. They can have a hierarchical organization, and they can be recursive. In contrast with some of the behavior-based approaches, I want the programs to be responsive to stored models of the environment as well as to their immediate sensory inputs.

The presentation of these ideas will be somewhat informal in line with my belief that formalization is best done after a certain amount of experience has been obtained. Although preliminary experiments indicate that the formalism works quite well, more work remains to be done to establish its place in agent control.

## 2. Teleo-Reactive Sequences

### 2.1 Condition-Action Rules

A teleo-reactive (T-R) sequence is an agent control program that directs the agent toward a goal (hence *teleo*) in a manner that takes into account changing environmental circumstances (hence *reactive*). In its simplest form, it consists of an ordered set of production rules:

$$\begin{array}{rcl} K_1 & \rightarrow & a_1 \\ K_2 & \rightarrow & a_2 \\ & \cdots & \\ K_i & \rightarrow & a_i \\ & \cdots & \\ K_m & \rightarrow & a_m \end{array}$$

The $K_i$ are conditions (on sensory inputs and on a model of the world), and the $a_i$ are actions (on the world or which change the model). A T-R sequence is interpreted in a manner roughly similar to the way in which some production systems are interpreted. The list of rules is scanned from the top for the first rule whose condition part is satisfied, and the corresponding action is executed. T-R sequences differ substantively from conventional production systems, however. T-R actions can be *durative* rather than discrete. A durative action is one that continues indefinitely. For example, a mobile robot is capable of executing the durative action *move*, which propels the robot ahead (say at constant speed) indefinitely. Such an action contrasts with a discrete one, such as *move forward one meter*. In a T-R sequence, a durative action continues so long as its corresponding condition remains the first true condition. When the first true condition changes, the action changes correspondingly. Thus, unlike production systems in computer science, the conditions must be *continuously* evaluated; the action associated with the *currently* first true condition is always the one being executed. An action terminates only when its energizing condition ceases to be the first true condition.

Indeed, rather than thinking of T-R sequences in terms of the computer science idea of discrete events, it is more appropriate to think of them as being implemented by circuitry. For example, the sequence above can be implemented by the circuit shown in figure 1. Furthermore, we imagine that the conditions, $K_i$, are also being continuously computed.

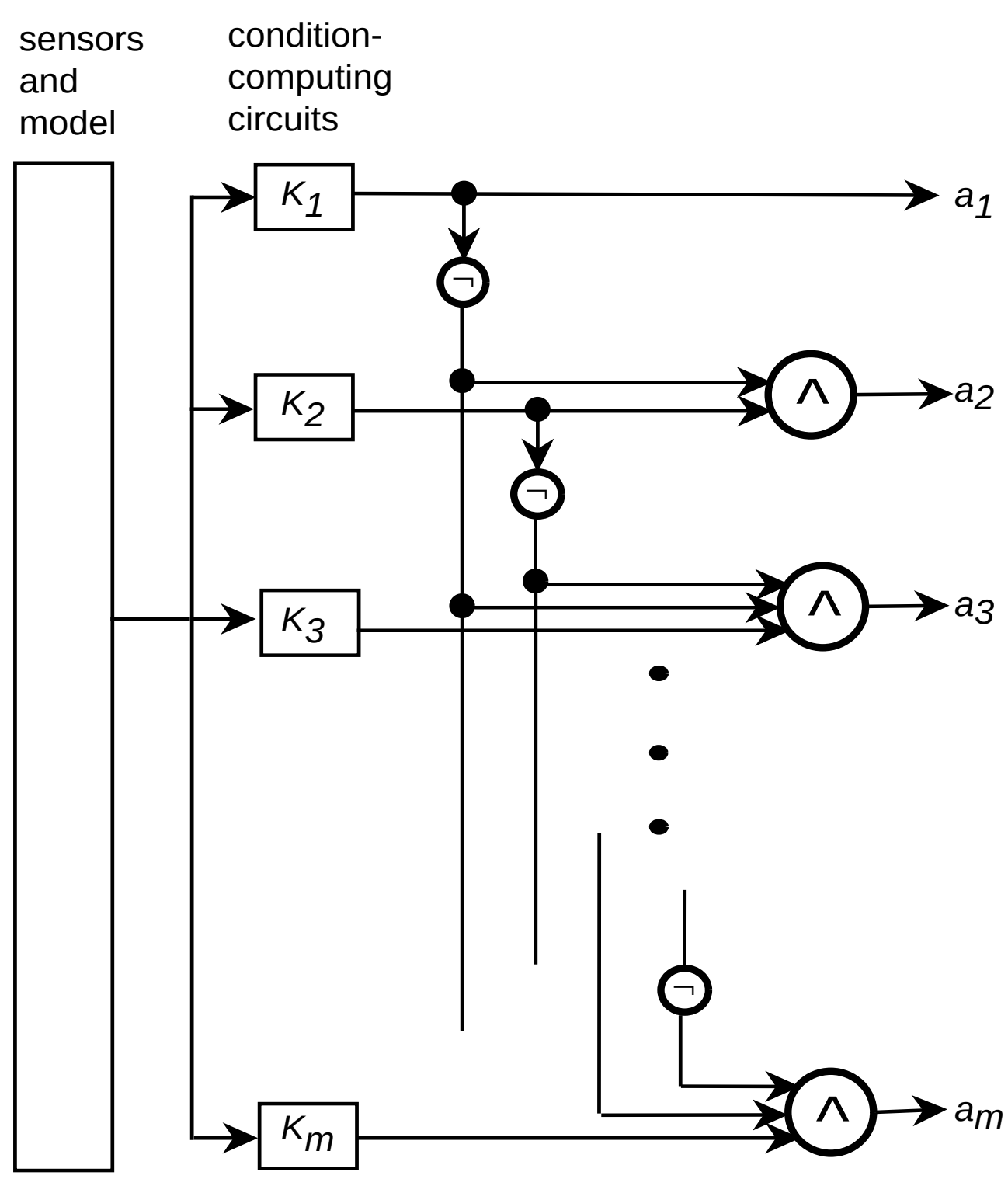

Figure 1: Implementing a T-R Sequence in Circuitry

The actions, $a_i$, of a T-R sequence can either be primitive actions, or they can be T-R sequences themselves. Thus, programs written in this formalism can be hierarchical (even recursive, as we shall see later). In the case of hierarchical programs, it is important to realize that *all* conditions at *all* levels of the hierarchy are continuously being evaluated; a high level sequence can redirect control through a different path of lower level sequences as dictated by the values of the conditions at the various levels.

In writing a T-R sequence, a programmer ordinarily works backward from whatever goal condition the sequence is being designed to achieve. The condition $K_1$ is taken to be the goal condition, and the corresponding action, $a_1$, is the null action. The condition $K_2$ is the weakest condition such that when it is satisfied (and $K_1$ is not), the durative execution of $a_2$ will (all other things being equal) eventually achieve $K_1$. And so on. Each non-null action, $a_i$, is supposed to achieve a condition, $K_j$, strictly higher in the list ($j < i$). The conditions are therefore *regressions* (Nilsson, 1980) of higher conditions through the actions that achieve those higher conditions.

Formally, we say that a T-R sequence satisfies the *regression property* if each condition, $K_i$ ($m \geq i > 1$), is the regression of some higher condition in the sequence, $K_j$ ($j < i$), through the action $a_i$. We say that a T-R sequence is *complete* if and only if $K_1 \vee \cdots \vee K_i \vee \cdots \vee K_m$ is a tautology. A T-R sequence is *universal* if it satisfies the regression property and is complete. It is easy to see that a universal T-R sequence will always achieve its goal condition, $K_1$, if there are no sensing or execution errors.

Sometimes an action does not have the effect that was anticipated by the agent's designer (the *normal* effect), and sometimes exogenous events (separate from the actions of the agent) change the world in unexpected ways. These phenomena, of course, are the reason continuous feedback is required. Universal T-R sequences, like universal plans (Schoppers, 1987), are robust in the face of occasional deviations from normal execution. They can also exploit serendipitous effects; it may accidentally happen that an action achieves a condition higher in the list of condition/action rules than normally expected. Even if an action sometimes does not achieve its normal effect (due to occasional sensing or execution errors), nevertheless *some* action will be executed. So long as the environment does not too often frustrate the achievement of the normal effects of actions, the goal condition of a universal T-R sequence will ultimately be achieved.

### 2.2 An Example

The following rather simple example should make these ideas more concrete. Consider the simulated robots in figure 2. Let's suppose that these robots can move bars around in their two-dimensional world. The robot on the right is holding a bar, and we want the other robot to go to and grab the bar marked $A$. We presume that this robot can sense its environment and can evaluate conditions which tell it whether or not it is already grabbing bar $A$ (*is-grabbing*), facing toward bar $A$ (*facing-bar*), positioned with respect to bar $A$ so that it can reach and grab it (*at-bar-center*), on the perpendicular bisector of bar $A$ (*on-bar-midline*), and facing a zone on the perpendicular bisector of bar $A$ from which it would be appropriate to move toward bar $A$ (*facing-midline-zone*). Let's assume also that these conditions have some appropriate amount of hysteresis so that hunting behavior is dampened. Suppose the robot is capable of executing the primitive actions *grab-bar*, *move*, and *rotate* with the obvious effects. Execution of the following T-R sequence will result in the robot grabbing bar $A$:

$$
\begin{array}{rcl}
\textit{is-grabbing} & \rightarrow & \textit{nil} \\
\textit{at-bar-center} \wedge \textit{facing-bar} & \rightarrow & \textit{grab-bar} \\
\textit{on-bar-midline} \wedge \textit{facing-bar} & \rightarrow & \textit{move} \\
\textit{on-bar-midline} & \rightarrow & \textit{rotate} \\
\textit{facing-midline-zone} & \rightarrow & \textit{move} \\
T & \rightarrow & \textit{rotate}
\end{array}
$$

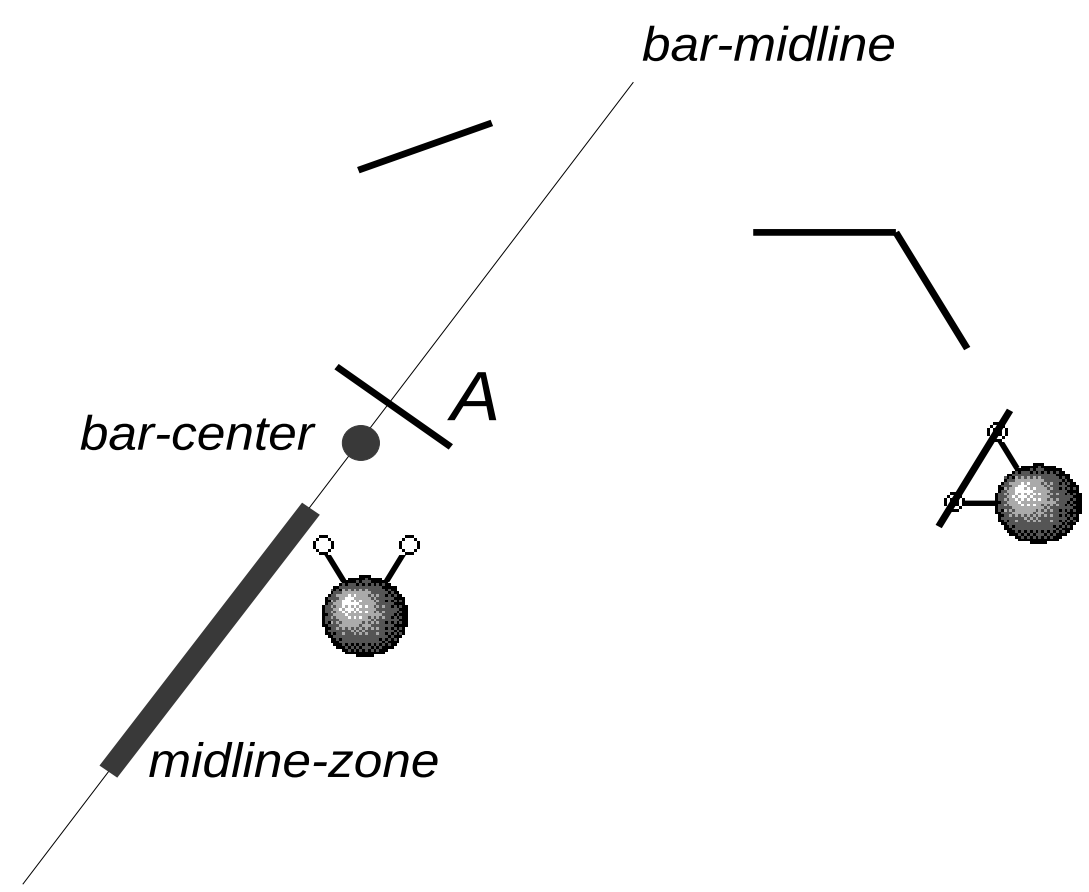

Figure 2: Robots and Bars

Notice how each properly executed action in this sequence achieves the condition in the rule above it. In this way, the actions inexorably proceed toward the goal. Occasional setbacks merely cause delays in achieving the goal so long as the actions usually[1] achieve their normal effects.

## 3. Teleo-Reactive Programs

### 3.1 Rules with Variables

We can generalize the notion of a T-R sequence by permitting the rules to contain free variables that are bound when the sequence is "called." We will call such a sequence a *T-R program.* Additional generality is obtained if we assume that the variables are not necessarily bound to constants but to quantities whose values are continuously being computed (as if by circuitry) as the environment changes.

A simple example involving having a robot go to a designated goal location in two dimensions will serve to illustrate. Suppose the goal location is given by the value of the variable *loc*. At run time, *loc* will be bound to a pair of $X, Y$ coordinates, although we allow the binding to change during run time. At any time during the process, the robot's $X, Y$ position is given by the value of the variable *position*. (We assume that the robot has some kind of navigational aid that reliably and continuously computes the value of *position*.) From the instantaneous values of *loc* and *position*, the robot can compute the direction that

1. We do not choose to define *usually* more precisely here, although a probabilistic analysis could be given.

it should face to proceed in a straight line toward *loc*. Let the value of this direction at any time be given by the value of the function *course*(*position*, *loc*). At any time during the process, the robot's angular heading is given by the value of the variable *heading*. Using these variables, the T-R program to drive the robot to *loc* is:

<u>*goto*(*loc*)</u>

$$\begin{array}{rcl} equal(position,\ loc) & \rightarrow & nil \\ equal(heading,\ course(position,\ loc)) & \rightarrow & move \\ T & \rightarrow & rotate \end{array}$$

Implementing *goto*(*loc*) in circuitry is straightforward. The single parameter of the program is *loc* whose (possibly changing) value is specified at run time by a user, by a higher level program, or by circuitry. The other (global) parameters, *position* and *heading*, are provided by circuitry, and we assume that the function *course* is continuously being computed by circuitry. Given the values of all of these parameters, computing which action to energize is then computed by circuitry in the manner of figure 1.

### 3.2 Hierarchical Programs

Our formalism allows writing hierarchical and recursive programs in which the actions in the rules are themselves T-R programs. As an example, we can write a recursive navigation program that calls *goto*. Our new navigation program requires some more complex sensory functions. Imagine a function *clear-path*(*place1*, *place2*) that has value $T$ if and only if the direct path is clear between *place1* and *place2*. (We assume the robot can compute this function, continuously, for *place1* = *position*, and *place2* equal to any target location.) Also imagine a function *new-point*(*place1*, *place2*) that computes an intermediate position between *place1* and *place2* whenever *clear-path* does not have value $T$. The value of *new-point* lies appropriately to the side of the obstacle determined to be between *place1* and *place2* (so that if the robot heads toward *new-point* first and then toward *place2*, it can navigate around the obstacle). Both *clear-path* and *new-point* are continuously computed by perceptual systems with which we endow the robot. We'll name our new navigation program *amble*(*loc*). Here is the code:

<u>*amble*(*loc*)</u>

$$\begin{array}{rcl} equal(position,\ loc) & \rightarrow & nil \\ clear\text{-}path(position,\ loc) & \rightarrow & goto(loc) \\ T & \rightarrow & amble(new\text{-}point(position,\ loc)) \end{array}$$

We show in figure 3 the path that a robot controlled by this program might take in navigating around the obstacles shown. (The program doesn't necessarily compute shortest paths; we present the program here simply as an illustration of recursion.) Note that if the obstacle positions or goal location change during execution, these changes will be reflected in the values of the parameters used by the program, and program execution will proceed in a manner appropriate to the changes. In particular, if a clear path ever becomes manifest between the robot and the goal location, the robot will abandon moving to any subgoals that it might have considered and begin moving directly to the goal.

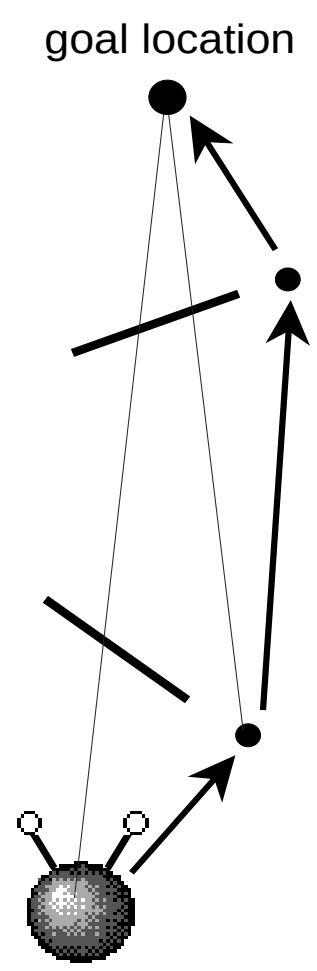

Figure 3: Navigating using *amble*

The continuous computation of parameters involved in T-R programs and the ability of high level programs to redirect control account for the great robustness of this formalism.

A formal syntax for T-R programs is given in (Nilsson, 1992).

### 3.3 Implementational Issues

The T-R formalism, with its implicit assumption of continuous computation of conditions and parameters, should be thought of as a fully legitimate "level" in the hierarchy of program structure controlling the agent, regardless of how this level is implemented by levels below—just as computer scientists think of list processing as a level of actual operation even though it is implemented by more primitive logical operations below. If we assume (as we do) that the pace of events in the agent's environment is slow compared with the amount of time taken to perform the "continuous" computations required in a T-R program, then the T-R programmer is justified in assuming "real" continuous sensing as s/he writes programs (even though the underlying implentation may involve discrete sampling). We recommend the T-R formalism only for those applications for which this assumption is justified. For those applications, the T-R level shields the programmer from having to worry about how that level is implemented and greatly facilitates program construction.

There are several different ways in which T-R programs can be interpreted into lower level implementations. It is beyond the scope of this paper to do more than point out some obvious methods, and we leave important questions about the properties of these methods to subsequent research. One method of implementation involves the construction of actual or simulated circuits according to the basic scheme of figure 1. First, the top level condition-computing circuits (including circuits for computing parameters used in the conditions) are constructed and allowed to function. A specific action, say $a_i$, is energized as a result. If $a_i$ is primitive, it is turned on, keeping the circuitry in place and functioning until some other top-level action is energized, and so on. If $a_i$ is a T-R sequence, the circuitry needed to implement it is constructed (just as was done at the top level), an action is selected, and so on—and all the while levels of circuitry above are left functioning. As new lower level circuitry is constructed, any circuitry no longer functioning (that is, circuitry no longer "called" by functioning higher level circuitry) can be garbage collected.

There are important questions of parameter passing and of timing in this process which I do not deal with here—relying on the assumption that the times needed to create circuitry and for the circuitry to function are negligible compared to the pace of events in the world. This assumption is similar to the *synchrony hypothesis* in the ESTEREL programming language (Berry & Gonthier, 1992) where it is assumed that a program's reaction "... takes no time *with respect to the external environment*, which remains invariant during [the reaction]."

Although there is no reason in principle that circuitry could not be simulated or actually constructed (using some sort of programmable network of logic gates), it is also straightforward to implement a T-R program using more standard computational techniques. T-R programs can be written as LISP `cond` statements, and durative actions can be simulated by iterating very short action increments. For example, the increment for the *move* action for a simulated robot might move the robot ahead by a small amount. After each action increment, the top level LISP `cond` is executed anew, and of course all of the functions and parameters that it contains are evaluated anew. In our simulations of robots moving in two-dimensional worlds (to be discussed below), the computations involved are sufficiently fast to effect a reasonable pace with apparent smooth motion.

This implementation method essentially involves sampling the environment at irregular intervals. Of course, there are questions concerning how the computation times (and thus the sampling rate) affect the real-time aspects of agent behavior which we do not address here—again assuming the sampling rate to be very short.

Whatever method is used to interpret T-R programs, care must be taken not to conflate the T-R level with the levels below. The programmer ought not to have to think about circuit simulators or sampling intervals but should imagine that sensing is done continuously and immediately.

### 3.4 Graphical Representations

The *goto* program can be represented by a graph as well as by the list of rules used earlier. The graphical representation of this program is shown in figure 4. The nodes are labeled by conditions, and the arcs by actions. To execute the graphical version of the program, we look for the shallowest true node (taking the goal condition as the root) and execute the action labeling the arc leading out from that node.

In the graph of figure 4, each action normally achieves the condition at the head of its arc (when the condition at the tail of the arc is the shallowest true condition). If there is more than one action that can achieve a condition, we would have a tree instead of a single-path graph. A more general graph, then, is a *teleo-reactive tree* such as that depicted in figure 5. T-R trees are executed by searching for the shallowest true node and executing the action labeling the arc leaving that node. Alternatively, we could search for that true node judged to be on a path of least cost to the goal, where some appropriate heuristic measure of cost is used. [For simplicity, the phrase "shallowest true node" will be taken to mean either the shallowest true node (literally) or the true node on a path of least cost to the goal.] Ties among several equally shallow true nodes are broken according to a fixed tie-breaking rule. In figure 5 we see that, in particular, there are at least two ways to achieve condition $K_1$. One way uses action $a_2$ (when $K_2$ is the shallowest true node), and one way uses action $a_3$ (when $K_3$ is the shallowest true node).

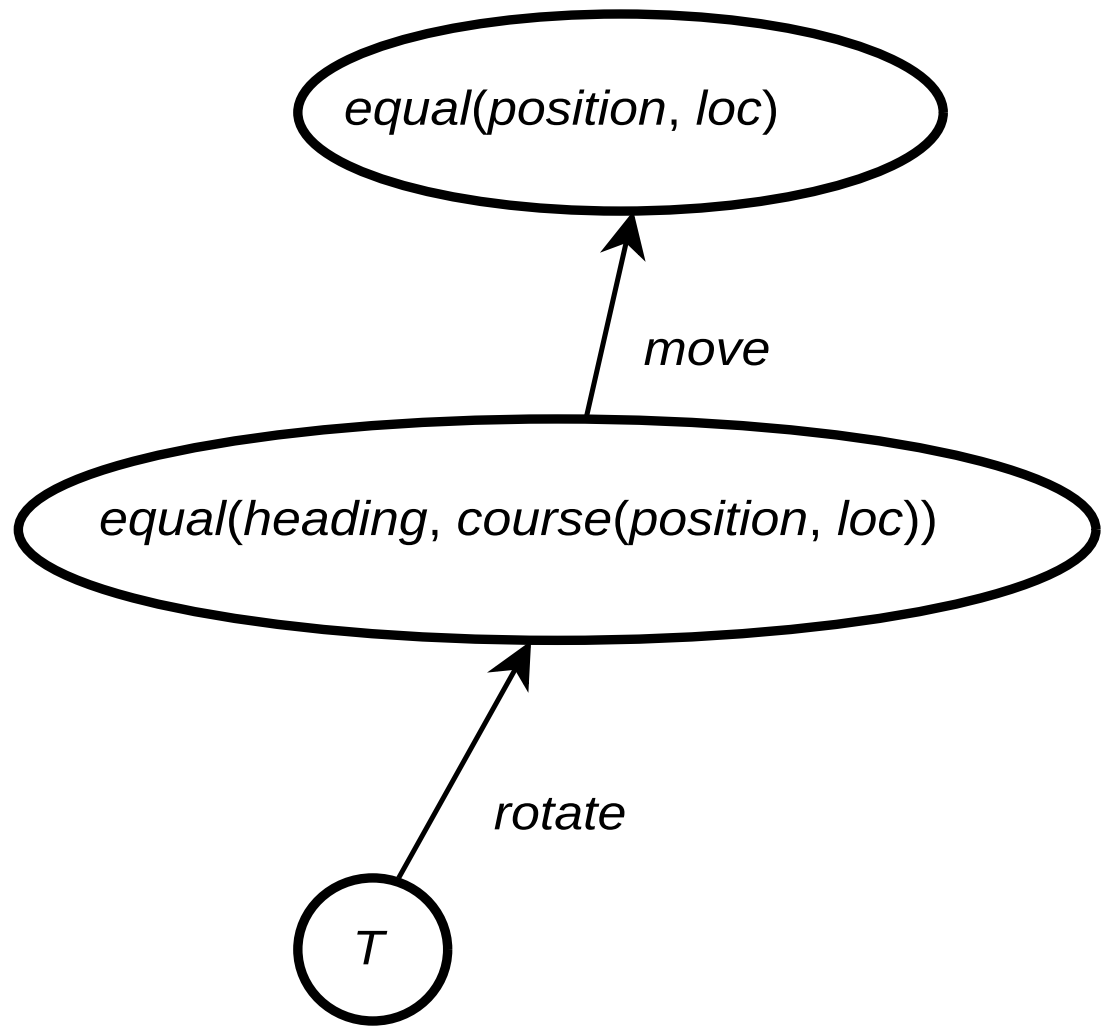

Figure 4: Graphical Representation of *goto*

In analogy with the definitions given for T-R sequences, a T-R tree *satisfies the regression property* if every non-root node is the regression of its parent node through the action linking it with its parent. A T-R tree is *complete* if the disjunction of all of its conditions is a tautology. A T-R tree is *universal* if and only if it satisfies the regression property and is also complete. With a fixed tie-breaking rule, a T-R tree becomes a T-R sequence. If a T-R tree is universal, then so will be the corresponding T-R sequence.

One might at first object to this method for executing a T-R tree on the grounds that the sequence of actions that emerge will hop erratically from one path to another. But if the tree satisfies the regression property, and if the heuristic for measuring cost to the goal is reasonable, then (however erratic the actions may appear to be), each successfully executed action brings the agent closer to the goal.

## 4. Experiments

We have carried out several preliminary experiments with agents programmed in this language (using LISP `cond` statements and short action increments). One set of experiments uses simulated robots acting in a two-dimensional space, called *Botworld* [2], of construction

2. The original Botworld interface, including the primitive perceptual functions and actions for its robots, was designed and implemented by Jonas Karlsson for the NeXT computer system (Karlsson, 1990). Sub-materials, structures made from these materials, and other robots. The construction materials are bars, and the robots are to build structures by connecting the bars in various ways. A robot can turn and move, can grab and release a suitably adjacent bar, can turn and move a grabbed bar, and can connect a bar to other bars or structures. The robots continuously sense whether or not they are holding a bar, and they "see" in front of them (giving them information about the location of bars and structures). Because of the existence of other robots which may change the world in sometimes unexpected ways, it is important for each robot to sense certain critical aspects of its environment continuously.

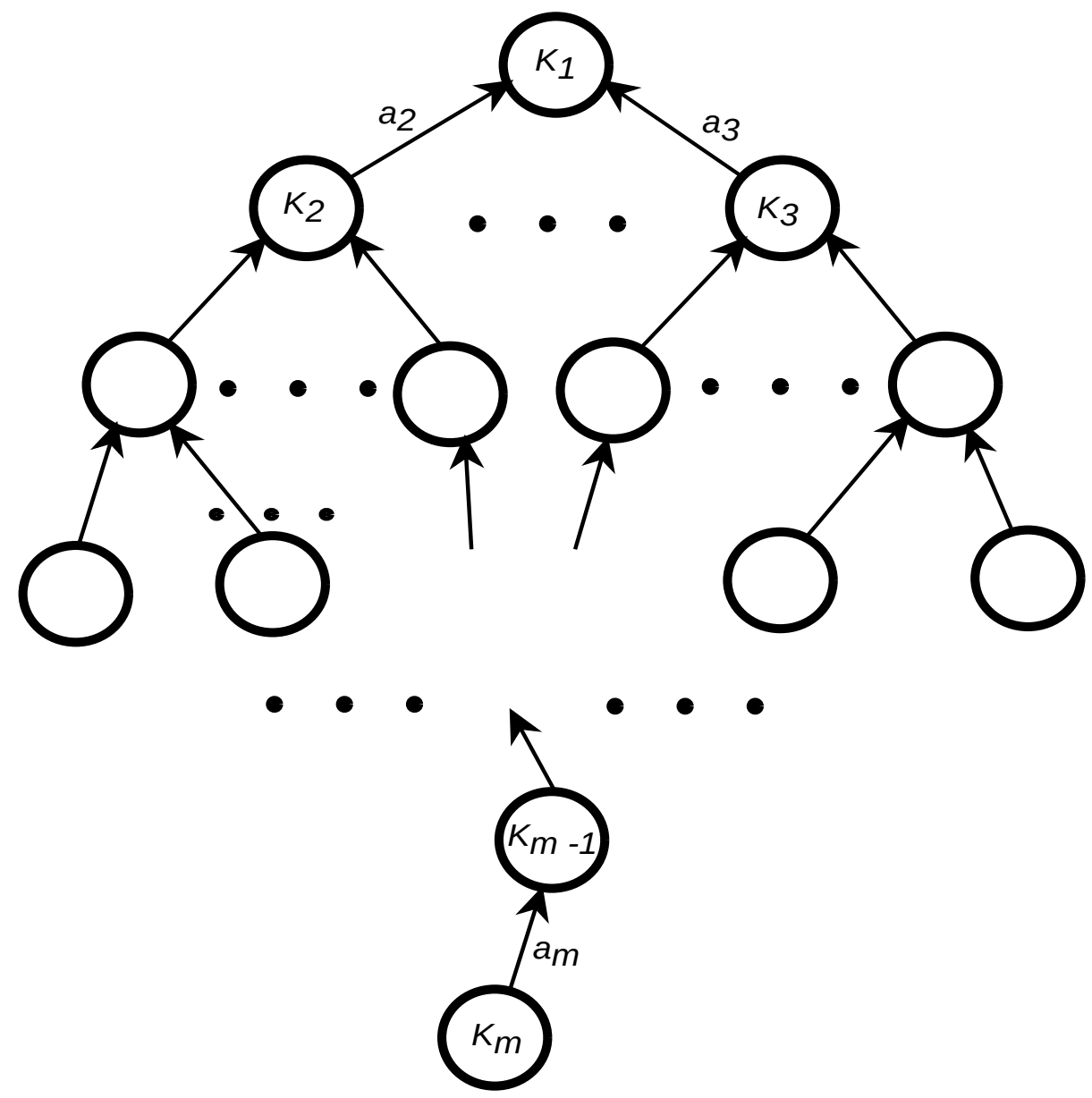

Figure 5: A T-R Tree

A typical Botworld graphical display is shown in figure 6.

We have written various T-R programs that cause the robots to build structures of various kinds (like the triangle being constructed in figure 6). A robot controlled by one of these programs exhibits homeostatic behavior. So long as the main goal (whatever it is) is satisfied, the robot is inactive. Whenever the goal (for whatever reason) is not satisfied, the robot becomes active and persists until it achieves the goal. If another agent achieves part or all of the goal, the robot carries on appropriately from the situation it finds itself in to complete the process.

In our experiments, the conditions used in the T-R rules are conditions on a model of the environment that the robot constructs from its sensory system and maintains separately from the T-R mechanism. The use of a model permits a robot to perform its actions in response to all the sensory stimuli (past and present) that have been used to help construct the model. But, if the T-R actions include *direct* changes to the model (in addition to those sequently, Patrick Teo implemented a version that runs under X-windows on any of several different workstations (Teo, 1991, 1992). The latter version allows the simulation of several robots simultaneously—each under the control of its own independently running process.

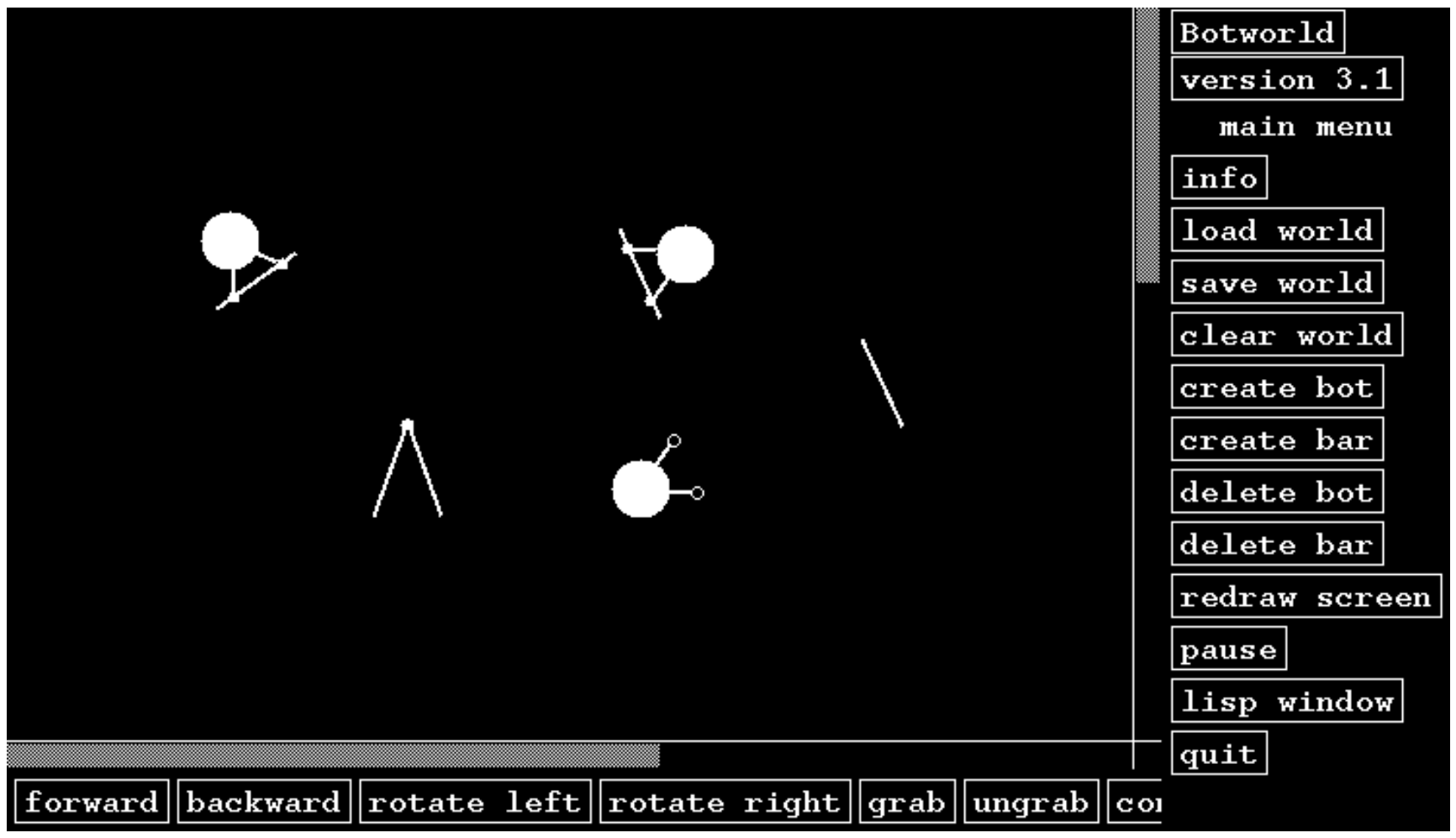

Figure 6: Botworld Display

changes resulting from perceived changes to the environment), then there is a potential for undesirable instabilities (as with any system with positive feedback). (The problem of how to model the environment and how this model should be updated in response to sensory data is a separate major research problem outside the scope of the work reported here.)

In other experiments, we have used the Nomadic Technologies 100 series mobile robot. The robot is equipped with a ring of 16 infrared sensors and a ring of 16 sonar sensors. It is controlled via a radio modem by a Macintosh II running Allegro Common Lisp. We have implemented robust T-R programs for some simple office-environment tasks, such as wall-following and corridor-following (Galles, 1993). The programs were initially developed and debugged using the Nomadics simulator of the actual robot; very few changes had to be made in porting the programs from the simulator to the robot. In performing these tasks, the robot is highly reactive and persistent even in the face of occasional extreme sonar or infrared range errors and deliberate attempts to confuse it. The robot quickly adapts to sudden changes in the environment, such as those caused by people sharing the hallways.

In writing T-R programs, one need only be concerned with inventing the appropriate predicates using the available perceptual functions and model database. One does not need to worry about providing interrupts of lower level programs so higher level ones can regain control. We have found that debugging T-R programs presents some challenges, though. Since they are designed to be quite robust in the face of environmental uncertainty, they also sometimes work rather well even though they are not completely debugged. These residual errors might not have undesirable effects until the programs are used in higher level programs—making the higher ones more difficult to debug.

## 5. Other Approaches for Specifying Behavior

There have been several formalisms proposed for prescribing sensory-directed, real-time activity in dynamic environments. Some of these are closely related to the T-R formalism proposed here. In this section I point out the major similarities and differences between T-R programs and a representative, though not complete, sample of their closest relatives. The other reactive formalisms are of two types, namely, those that sample their environments at discrete intervals (perhaps rapidly enough to be sufficiently reactive), and those that create circuitry (like T-R programs). The discrete-sampling systems do not abstract this activity into a higher level in which the environment is monitored continuously, and most of the circuitry-creating systems do so prior to run time (unlike T-R programs which create circuitry at run time).

### 5.1 Discrete-Sampling Systems

#### 5.1.1 Production Systems

As has already been mentioned, T-R programs are similar to production systems (Waterman & Hayes-Roth, 1978). The intermediate-level actions (ILAs) used in the SRI robot Shakey (Nilsson, 1984) were programmed using production rules and were very much like T-R programs. A T-R program also resembles a plan represented in triangle-table form constructed by STRIPS (Fikes, Hart & Nilsson, 1972). Each of the conditions of a T-R sequence corresponds to a triangle table kernel. In the PLANEX execution system for triangle tables, the action corresponding to the highest-numbered satisfied kernel is executed. A major difference between all of these previous production-system style programs and T-R programs is that T-R programs are continuously responsive to the environment while ordinary production systems are not.

#### 5.1.2 Reactive Plans

Several researchers have adopted the approach of using the current situation to index into a set of pre-arranged action sequences (Georgeff & Lansky, 1987; Schoppers, 1987; Firby, 1987). This set can either be large enough to cover a substantial number of the situations in which an agent is likely to find itself or it can cover all possible situations. In the latter case, the plan set is said to be *universal*. Unlike T-R programs, these systems explicitly sample their environments at discrete time steps rather than continuously. As with T-R programs, time-space trade-offs must be taken into account when considering how many different conditions must be anticipated in providing reactive plans. Ginsberg has noted that in several domains, the number of situations likely to be encountered by the agent is so intractably large that the agent is forced to postpone most of its planning until run time when situations are actually encountered (Ginsberg, 1989). (For further discussion of this point, see (Selman, 1993).) T-R programs have the advantage that at least a rudimentary form of planning, namely parameter binding, is done at run time. The PRS system (Georgeff & Lansky, 1987) is capable of more extensive planning at run time as well as reacting appropriately to its current situation.

#### 5.1.3 Situated Control Rules

Drummond (Drummond, 1989) introduces the notion of a *plan net* which is a kind of Petri net (Reisig, 1985) for representing the effects of actions (which can be executed in parallel). Taking into account the possible interactions of actions, he then projects the effects of all possible actions from a present state up to some horizon. These effects are represented in a structure called a *plan projection*. The plan projection is analyzed to see, for each state in it, which states *possibly* have a path to the goal state. This analysis is a forward version of the backward analysis used by a programmer in producing a T-R tree. *Situated control rules* are the result of this analysis; they constrain the actions that might be taken at any state to those which will result in a state that still possibly has a path to the goal. Plan nets and Petri nets are based on discrete events and thus are not continuously responsive to their environments in the way that T-R programs are.

### 5.2 Circuit-Based Systems

Kaelbling has proposed a formalism called GAPPS (Kaelbling, 1988; Kaelbling & Rosenschein, 1990), involving *goal reduction rules*, for implicitly describing how to achieve goals. The GAPPS programmer defines the activity of an agent by providing sufficient goal reduction rules to connect the agent's goals with the situations in which it might find itself. These rules are then compiled into circuitry for real-time control of the agent. Rosenschein and Kaelbling (Rosenschein & Kaelbling, 1986) call such circuitry *situated automata*.

A collection of GAPPS rules for achieving a goal can be thought of as an implicit specification of a T-R program in which the computations needed to construct the program are performed when the rules are compiled. The GAPPS programmer typically exerts less specific control over the agent's activity—leaving some of the work to the search process performed by the GAPPS compiler. For example, a T-R program to achieve a goal, $p$ , can be implicitly specified by the following GAPPS rule:

```
(defgoalr (ach ?p)
              (if ((holds ?p) (do nil))
              ((holds (regress ?a ?p)) (do ?a))
              (T ach (regress ?a ?p)) ))
```

The recursion defined by this rule bottoms out in rules of the form:

```
(defgoalr (ach φ)
             ((holds ψ) (do α)) )
```

where $\phi$ and $\psi$ are conditions and $\alpha$ is a specific action.

GAPPS compiles its rules into circuitry before run time, whereas the circuit implementation of a T-R program depends on parameters that are bound at run time. Both systems result in control that is continuously responsive to the environment.

In implementing a system to play a video game, Chapman (Chapman, 1990) compiles production-like rules into digital circuitry for real-time control using an approach that he calls "arbitration macrology." As in situated automata, the compilation process occurs prior to run time.

Brooks has developed a behavior language, BL, (Brooks, 1989), for writing reactive robot control programs based on his "subsumption architecture" (Brooks, 1986). A similar language, ALFA, has been implemented by Gat (Gat, 1991). Programs written in these languages compile into structures very much like circuits. Again, compilation occurs prior to run time. It has been relatively straightforward to translate examples of subsumption-architecture programs into T-R programs.

In all of these circuit-based systems, pre-run-time compiling means that more circuitry must be built than might be needed in any given run because all possible contingencies must be anticipated at compile time.[3] But in T-R programs, parameters are bound at run time, and only that circuitry required for these specific bindings is constructed.

## 6. Future Work

The T-R formalism might easily be augmented to embody some features that have not been discussed in this paper. Explicit reference to time in specifying actions might be necessary. For example, we might want to make sure that some action $a$ is not initiated until after some time $t_1$ and ceases after some time $t_2$. Time predicates, whose time terms are evaluated using an internal clock, may suffice for this purpose.

Also, in some applications we may want to control which conditions in a T-R program are actually tested. It may be, for example, that some conditions won't have to be checked because their truth or falsity can be guessed with compelling accuracy.

Simultaneous and asynchronous execution of multiple actions can be achieved by allowing the right-hand side of rules to contain *sets* of actions. Each member of the set is then duratively executed asynchronously and independently (so long as the condition in the rule that sustains this set remains the highest true condition). Of course, the programmer must decide under what conditions it is appropriate to call for parallel actions. Future work on related formalisms might reveal ways in which parallel actions might *emerge* from the interaction of the program and its environment rather than having to be explicitly programmed.

Although we intend that T-R programs for agent control be written by human programmers, we are also interested in methods for modifying them by automatic planning and machine learning. We will briefly discuss some of our preliminary ideas on planning and learning here.

T-R trees resemble the search trees constructed by those planning systems that work backwards from a goal condition. The overall goal is the root of the tree; any non-root node $g_i$ is the regression of its parent node, $g_j$ through the action, $a_k$, connecting them. This similarity suggests that T-R trees can be constructed (and modified) by an automatic planning system capable of regressing conditions through durative actions. Indeed triangle tables (Fikes, Hart & Nilsson, 1972), a degenerate form of T-R tree consisting of only a single path, were constructed by an automatic planning system and an EBL-style generalizer (Mitchell, Keller & Kedar-Cabelli, 1986).

The reader might object that there is no reason to suppose that the search trees produced by an automatic planning process will contain nodes whose conditions are those that the agent is likely to encounter in its behavior. A process of incremental modification, however, should gradually make these constructed trees more and more matched to the agent's environment. If a tree for achieving a desired goal has no true nodes in a certain situation,

3. Agre's "running arguments" construct (Agre, 1989) is one example of a circuit-based system that can add circuitry at run time as needed.

it is as if the search process employed by an automatic planner had not yet terminated because no subgoal in the search tree was satisfied in the current state. In this case, the planning system can be called upon to continue to search; that is, the existing T-R tree will be expanded until a true node is produced. Pruning of T-R trees can be accomplished by keeping statistics on how often their nodes are satisfied. Portions of the trees that are never or seldom used can be erased. Early unpublished work by Scott Benson indicates that T-R programs can be effectively generated by automatic planning methods (Benson, 1993).

In considering learning mechanisms, we note first that T-R sequences are related to a class of Boolean functions that Rivest has termed *k-decision lists* (Rivest, 1987; Kohavi & Benson, 1993). A $k$-decision list is an ordered list of condition-value pairs in which each condition is a conjunction of Boolean variables of length at most $k$, and each value is a truth value ($T$ or $F$). The value of the Boolean function represented by a $k$-decision list is that value associated with the highest true condition. Rivest has shown that such functions are polynomially PAC learnable and has presented a supervised learning procedure for them. We can see that a T-R sequence whose conditions are limited to $k$-length conjunctions of Boolean features is a slight generalization of $k$-decision lists. The only difference is that such a T-R sequence can have more than two different "values" (that is, actions). We observe that such a T-R sequence (with, say, $n$ different actions) is also PAC learnable since its actions can be encoded with $\log_2 n$ decision lists. George John (John, 1993) has investigated a supervised learning mechanism for learning T-R sequences.

Typically, the conditions used in T-R programs are conjunctions of propositional features of the robot's world and/or model. Because a linear threshold function can implement conjunctions, one is led to propose a neural net implementation of a T-R sequence. The neural net implementation, in turn, evokes ideas about possible learning mechanisms. Consider the T-R sequence:

$$
\begin{array}{rcl}
K_1 & \rightarrow & a_1 \\
K_2 & \rightarrow & a_2 \\
 & \cdots & \\
K_i & \rightarrow & a_i \\
 & \cdots & \\
K_m & \rightarrow & a_m
\end{array}
$$

Suppose we stipulate that the $K_i$ are linear threshold functions of a set of propositional features. The $a_i$ are not all necessarily distinct; in fact we will assume that there are only $k \leq m$ distinct actions. Let these be denoted by $b_1, \cdots, b_k$. The network structure in figure 7 implements such a T-R sequence.

The propositional features tested by the conditions are grouped into an $n$-dimensional binary (0,1) vector, $\mathsf{X}$ called the *input vector*. The $m$ conditions are implemented by $m$ threshold elements having weighted connections to the components of the input vector. The process of finding the first true condition is implemented by a layer containing appropriate inhibitory weights and AND units such that only one AND unit can ever have an output value of 1, and that unit corresponds to the first true condition. A unique action is associated with each condition through a layer of binary-valued weights and OR-unit *associators*. Each AND unit is connected to one and only one associator by a non-zero weight. Since only one AND unit can have a non-zero output, only that unit's associator can have a non-zero output. (But each associator could be connected to multiple AND units.) For example, if action $b_i$ is to be associated with conditions $K_j$ and $K_l$, then there will be unit weights from the $j$-th and $l$-th AND units to the associator representing action $b_i$ and zero-valued weights from all other AND units to that associator. The action selected for execution is the action corresponding to the single associator having the non-zero output. We are investigating various learning methods suggested by this neural net implementation.

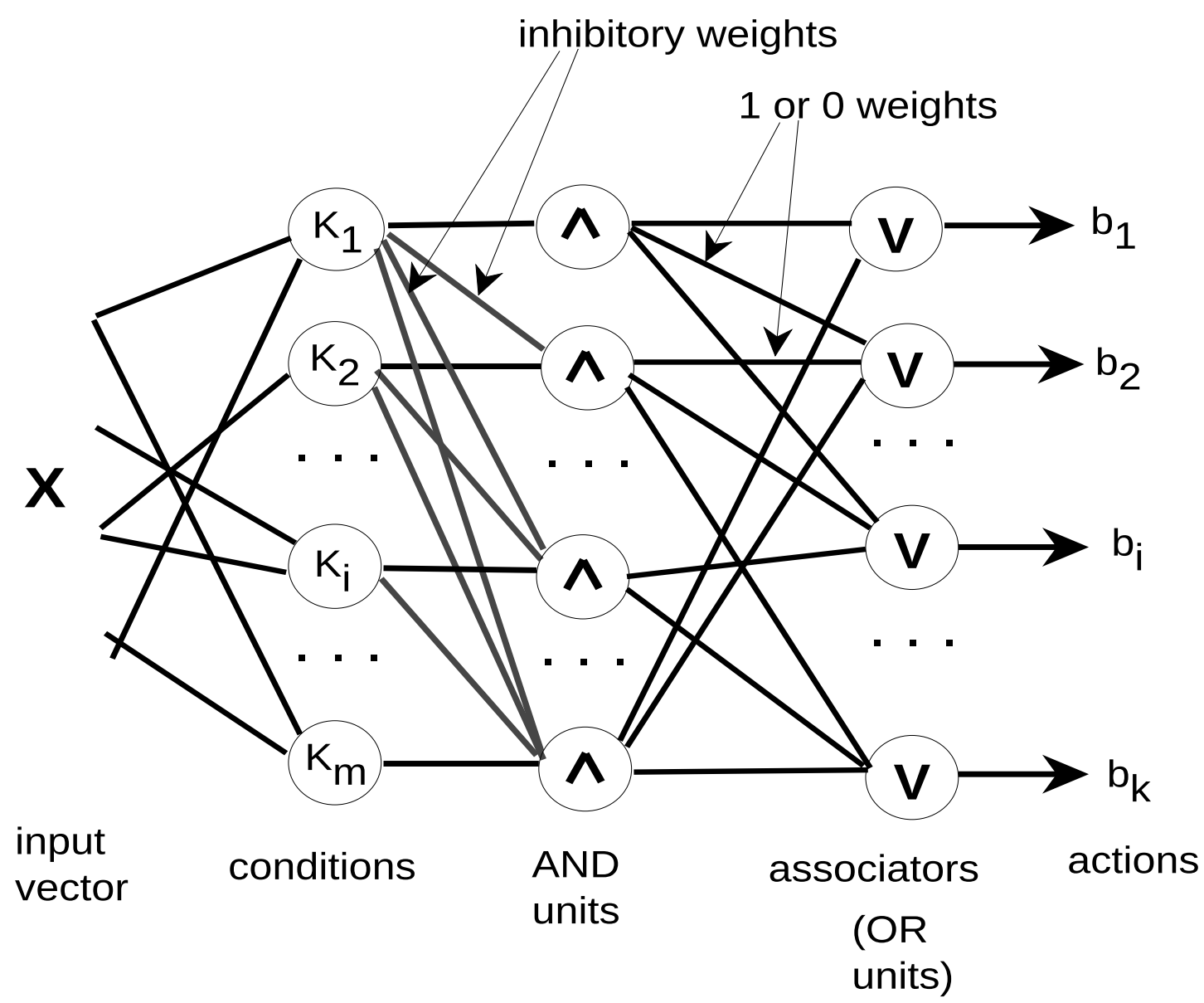

Figure 7: A Neural Net that Implements a T-R Sequence

Work must also be done on the question of what constitutes a *goal*. I have assumed goals of achievement. Can mechanisms be found that continously avoid making certain conditions true (or false) while attempting to achieve others? Or suppose priorities on a number of possibly mutually contradictory conditions are specified; what are reasonable methods for attending to those achievable goals having the highest priorities?

Also, it will be interesting to ask in what sense T-R programs can be proved to be correct. It would seem that verification would have to make assumptions about the dynamics of the environment; some environments might be so malevolent that agents in them could never achieve their goals. Even so, a verifier equipped with a model of the effects of actions could at least check to see that the regression property was satisfied and note any lapses.

More work remains on methods of implementing or interpreting T-R programs and the real-time properties of implementations. These properties will, of course, depend on the depth of the T-R program hierarchy and on the conditions and features that must be evaluated.

Finally, it might be worthwhile to investigate "fuzzy" versions of T-R trees. One could imagine fuzzy predicates that would energize actions with a "strength" that depends on the degree to which the predicates are true. The SRI robot, Flakey, uses a fuzzy controller (Saffiotti, Ruspini & Konolige, 1993).

## 7. Conclusions

I have presented a formalism for specifying actions in dynamic and uncertain domains. Since this work rests on ideas somewhat different than those of conventional computer science, I expect that considerably more analysis and experimentation will be required before the T-R formalism can be fully evaluated. The need in robotics for control-theoretic ideas such as homeostasis, continuous feedback, and stability appears to be sufficiently strong, however, that it seems appropriate for candidate formalisms embodying these ideas to be put forward for consideration.

Experiments with the language will produce a stock of advice about how to write T-R programs effectively. Already, for example, it is apparent that a sustaining condition in a T-R sequence must be carefully specified so that it is no more restrictive than it really needs to be; an overly restrictive condition is likely to be rendered false by the very action that it is supposed to sustain before that action succeeds in making a higher condition in the sequence true. But, of course, overly restrictive conditions won't occur in T-R programs that satisfy the regression property.

To be usefully employed, T-R programs (or any programs controlling agent action) need to be embodied in an overall agent architecture that integrates perceptual processing, goal selection, action computation, environmental modeling, and planning and learning mechanisms. Several architectural schemes have been suggested, and we will not summarize them here except to say that three layers of control are often delineated. A typical example is the SSS architecture of Connell (Connell, 1993). His top (Symbolic) layer does overall goal setting and sequencing, the middle (Subsumption) level selects specific actions, and the lower (Servo) level exerts standard feedback control over the effectors. We believe T-R programs would most appropriately be used in the middle level of such architectures.

The major limitation of T-R programs is that they involve much more computation than do programs that check only relevant conditions. Most of the conditions computed by a T-R program in selecting an action are either irrelevant to the situation at hand or have values that might be accurately predicted (if the programmer wanted to take the trouble to do so). We are essentially trading computing time for ease of programming, and our particular trade will only be advantageous in certain applications. Among these, I think, is the mid-level control of robots and (possibly) software agents.

In conclusion, there are three main features embodied in the T-R formalism. One is *continuous computation* of the parameters and conditions on which action is based. T-R programs allow for continuous feedback while still supporting parameter binding and recursion. The second feature is the *regression* relationship between conditions in a T-R program. Each condition is the regression of some condition closer to the goal through an action that normally achieves that closer-to-the-goal condition. The regression property assures robust goal-seeking behavior. Third, the conceptual circuitry controlling action is constructed at *run time*, and this feature permits programs to be universal while still being compact. In addition, T-R programs are intuitive and easy to write and are written in a formalism that is compatible with automatic planning and learning methods.

## Acknowledgements

I trace my interest in reactive, yet purposive, systems to my early collaborative work on triangle tables and ILAs. Several former Stanford students, including Jonas Karlsson, Eric Ly, Rebecca Moore, and Mark Torrance, helped in the early stages of this work. I also want to thank my sabbatical hosts, Prof. Rodney Brooks at MIT, Prof. Barbara Grosz at Harvard, and the people at the Santa Fe Institute. More recently, I have benefitted from discussions with Scott Benson, George John, and Ron Kohavi. I also thank the anonymous referees for their helpful suggestions. This work was performed under NASA Grant NCC2-494 and NSF Grant IRI-9116399.